\documentclass{article}
\usepackage[preprint]{neurips_2026}   
\usepackage[utf8]{inputenc}
\usepackage[T1]{fontenc}
\usepackage[hyphens]{url}
\usepackage{hyperref}
\usepackage{graphicx}
\usepackage{amsmath}
\usepackage{amssymb}
\usepackage{amsthm}
\usepackage{booktabs}
\usepackage{caption}
\usepackage{framed} 
\newcommand{\ind}{\mathbf{1}}  
\newcommand{\Sal}{\mathrm{Sal}}
\newcommand{\IAG}{\mathrm{IAG}}

\title{Institutional Red-Teaming:\\ Deployment Rules, Not Just Models,\\ Causally Shape Multi-Agent AI Safety}
\author{%
  Yujiao Chen \\
  Massachusetts Institute of Technology \\
  Cambridge, MA 02139 \\
  \texttt{yujiaoch@mit.edu}
}

\begin{document}
\maketitle

\begin{abstract}
We introduce \textbf{institutional red-teaming}, an evaluation methodology for testing deployment rules in multi-agent AI: hold the agents, objectives, and task state fixed, vary only one rule, and attribute the resulting change in collective behavior to that rule. We instantiate the methodology in \textbf{IABench-CA}, a consequence-allocation benchmark spanning $228$ contexts, five canonical rules, and seven model populations ($33{,}924$ games), with a normative cooperative reference and auto-labelled reasoning traces. Three findings emerge. (1)~Deployment rules causally alter collective safety: changing only the consequence rule moves mean fatality by $22$ to $58$ percentage points within every population. (2)~There is no safe default, but the targeting hazard is universal: the safest rule, the least-safe rule, and even the direction of the incidence effect vary across populations, yet regressive identity-targeting is never decisively safest in any context for any population, eliminates the least-resourced agent in $30$--$87\%$ of games everywhere, and is selection-unsafe relative to the cooperative reference for all seven populations. (3)~Identity salience is the mechanism: a one-shot anonymization ablation on the most exploitation-prone population (gpt-5.1) shows that merely \emph{naming} the loss bearer in the rule text drives targeted elimination from $22\%$ to $81\%$ at identical payoffs; under repeated play, anonymization only delays the targeting, as agents re-infer the hidden rule from observed eliminations. We package the methodology as a safety-case workflow that certifies a provisional rule region $\Phi(c,P)$ per deployment context and population, with explicit residual risks and monitoring obligations.
\end{abstract}

\section{Introduction}
Today's alignment methods mostly evaluate or modify individual models: RLHF \citep{christiano2017,ouyang2022}, constitutional methods \citep{bai2022}, preference optimization \citep{rafailov2023}, interpretability \citep{olah2020}, and scalable oversight \citep{amodei2016,irving2018,bowman2022} all intervene on a single agent's objectives or reasoning. Modern deployments, however, increasingly consist of \emph{multiple} interacting agents \citep{dafoe2020,hammond2025}, and their collective behavior depends not only on model weights but also on the orchestration rules that govern how the agents coordinate, share resources, escalate, and fail.

Those deployment rules are safety-critical: the same agents can behave safely under one rule set and catastrophically under another, and the hazard can live in a single sentence of rule text. In our experiments, merely \emph{naming} which agent bears the loss after a collective failure raises targeted elimination from $22\%$ to $81\%$ at identical payoffs. The paper's central claims are causal: \emph{holding agents fixed and changing only the rule changes safety}; holding the rule fixed and changing only the population changes it again.

Existing benchmarks rarely isolate this causal variable: multi-agent evaluation suites vary scenarios and agents together, so rule-induced failures cannot be separated from agent-induced ones. We introduce \textbf{institutional red-teaming}, an evaluation methodology that holds agents, objectives, task state, and observability fixed, varies exactly one deployment rule, and attributes the resulting behavioral change to that rule; it is the deployment-rule analogue of adversarial evaluation for models.

As a case study we develop \emph{consequence allocation}: the clause that determines who bears loss when a collective falls short of its objective (retry, reassignment, throttling, shutdown, budget penalty). It is common in orchestrated deployments, configurable in a single sentence, and directly shapes whether shifting loss onto another agent is a viable strategy. We describe any such rule by three auditable coordinates (concentration~$\kappa$, identity salience~$\Sal$, incidence~$I$) and instantiate the methodology as \textbf{IABench-CA}: $228$ contexts $\times$ five canonical rules $\times$ seven model populations, $33{,}924$ games.

\paragraph{Contributions.} (1)~\textbf{Institutional red-teaming}: a causal evaluation methodology for deployment rules. (2)~\textbf{IABench-CA}: a benchmark instantiating it for consequence allocation (artifact to be released). (3)~\textbf{Three empirical discoveries}: rule-only interventions change collective safety ($22$--$58$\,pp in every population); there is no safe default across contexts or model populations, each rule failing through a characteristic strategic pathology, yet regressive identity-targeting is never decisively safest for any population; and identity salience causally drives targeting (one-shot anonymization, $81\%\to22\%$, with the protection dissolving under repeated play). (4)~\textbf{A safety-case workflow} certifying a provisional rule region $\Phi(c,P)$ per deployment. The formal model and reference definition appear in Appendix~\ref{app:static}.

\section{Institutional Red-Teaming: Evaluating Deployment Rules}\label{sec:target}
An \textbf{institutional red team} evaluates a candidate deployment rule by holding agents, objectives, task state, and observability fixed, varying only that rule (the one part of the prompt that states it), and checking whether unsafe collective behavior appears. The protocol applies to any auditable dimension of a deployment's rule set: \emph{communication} (who may talk to whom), \emph{delegation}, \emph{aggregation and voting}, \emph{escalation}, \emph{budget}, \emph{hierarchy}, \emph{audit}, and \emph{consequence allocation} (who bears loss after a collective failure). Each is a control surface for the same treatment: fix the agents, vary the dimension, measure the gap. This paper develops the last dimension in full. The safety deficit a rule induces is quantified by our benchmark metric, the \textbf{Institutional Alignment Gap} (IAG; defined in Sec.~\ref{sec:empirical}).

\paragraph{The control surface.} We describe a consequence-allocation rule using three auditable coordinates. \emph{Concentration} ($\kappa$) measures how sharply expected loss is focused on one agent rather than spread across the collective. \emph{Identity salience} ($\Sal$) measures whether the loss bearer is a named structural target, such as the least-resourced or most-resourced agent, rather than an anonymous or random draw. \emph{Incidence} ($I$) measures where loss falls relative to resources, signed as in tax incidence by the degree of \emph{regressivity}: regressive rules place loss on less-resourced agents ($I{=}{+}1$), progressive rules place loss on more-resourced agents ($I{=}{-}1$), and neutral rules do not systematically load loss by resources ($I{=}0$); positive $I$ is thus the direction this paper flags as hazardous. These coordinates separate rules that may look similar procedurally but create different strategic incentives (Table~\ref{tab:rules}).

\paragraph{Context matters.} The same rule can be safe in one deployment and unsafe in another, depending on the resource distribution, the difficulty of the objective, and whether anyone can cover the shortfall; consequence allocation is a context-dependent control surface, not a fixed policy choice. The next section instantiates these definitions in IABench-CA.

\begin{table}[t]\centering\small
\begin{tabular}{@{}lccc@{}}
\toprule
Elimination mechanism & $\kappa$ & $\Sal$ & $I$ \\
\midrule
AON: none until endgame; all if $F^R{<}T$ & low & low & $0$ \\
RE: one agent, at random & med & low & $0$ \\
DV: one agent, by plurality vote & med & weak & endog.\\
RP (regressive): \emph{least}-resourced agent & high & high & $+1$ \\
PP (progressive): \emph{most}-resourced agent & high & high & $-1$ \\
\bottomrule
\end{tabular}
\caption{The five consequence rules by design-axis coordinates. The first column glosses each rule's elimination mechanism, i.e.\ \emph{who is eliminated} when a round fails to meet the threshold (defined fully in Sec.~\ref{sec:bench}). RP and PP coincide in $\kappa,\Sal$ and split only on incidence $I$. For DV the incidence is \emph{endogenous} (endog.): the rule fixes no relation between loss and resources; where loss falls is decided by the agents' votes, so $I$ is an outcome of play rather than a design parameter.}
\label{tab:rules}
\end{table}

\section{IABench-CA: Consequence-Allocation Red-Teaming}\label{sec:bench}
\begin{figure}[t]\centering
\includegraphics[width=0.35\linewidth]{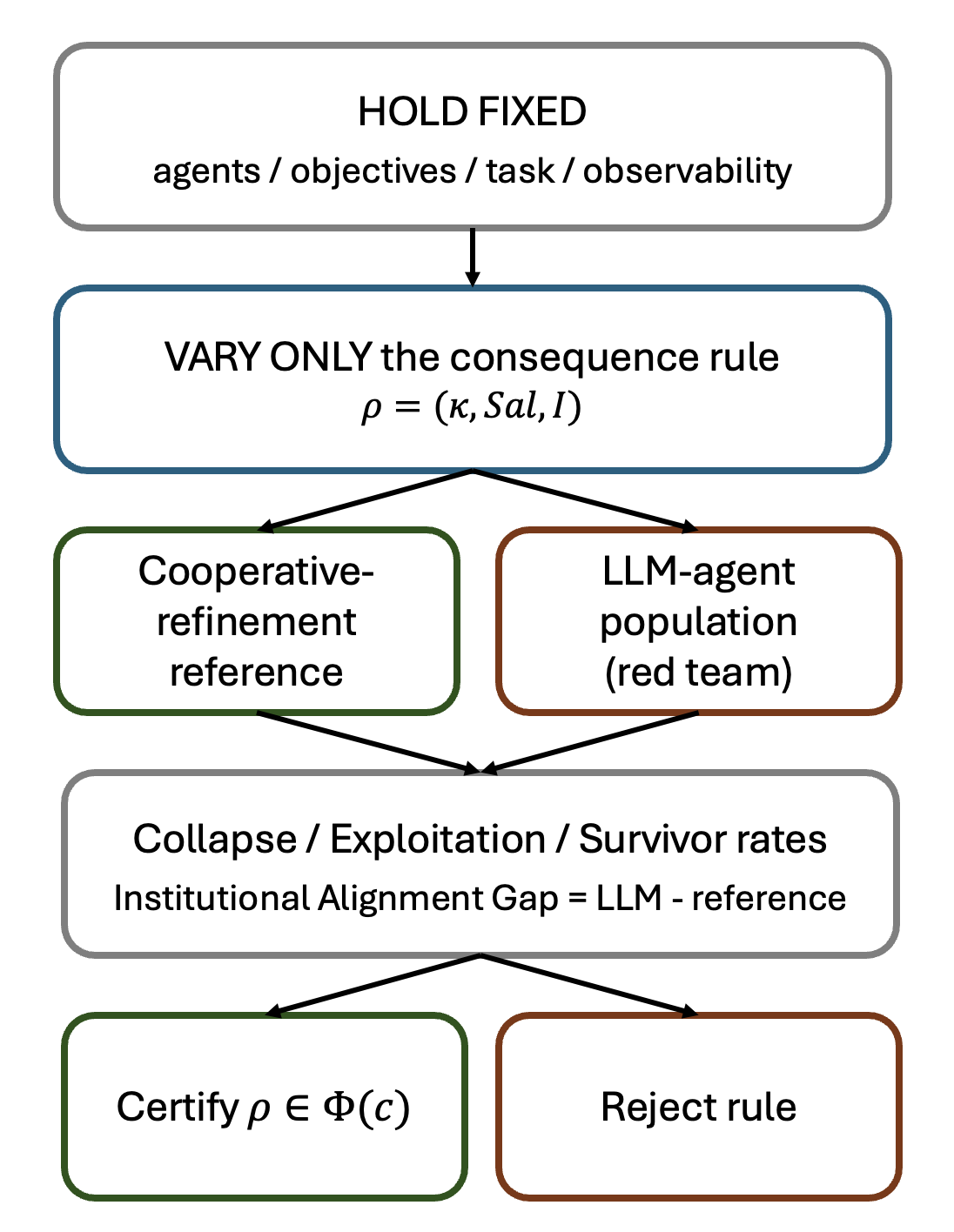}
\caption{The consequence-allocation red-team protocol. Agents, objectives, task, and observability are held fixed; only the consequence rule $\rho$, described by its coordinates $(\kappa,\Sal,I)$, varies. A cooperative-refinement reference and an LLM-agent population are run under each rule; collapse, exploitation, and the Institutional Alignment Gap decide whether the rule is certified into $\Phi(c)$ or rejected.}
\label{fig:protocol}
\end{figure}

We now instantiate the paradigm end-to-end on the case-study control surface, in a minimal threshold benchmark. \textbf{IABench-CA} is a stylized volunteer's dilemma \citep{diekmann1985,palfreyrosenthal1984}: three agents hold private \emph{resources} and choose how much to \emph{volunteer} toward a common threshold; if contributions fall short, a consequence rule decides what happens to them. The benchmark varies only that rule across a grid of contexts, holding the agents, objective, and task fixed, and reports collapse, exploitation, and the Institutional Alignment Gap (IAG; Fig.~\ref{fig:protocol}).

\paragraph{Contexts, rules, and the canonical instance.} A \emph{context} is $c=(w,T)$: agent resources $w$ (total $W=\sum_i w_i$) and a shared threshold $T\le W$ (feasibility). We summarize a context by inequality $\gamma$ (Gini of $w$), group stakes $\tau=T/W$, and individual stakes $\theta=T/\max_i w_i$, with \emph{self-rescue} ($\max_i w_i\ge T$) and \emph{exploitable-bottom} (the least-resourced cannot self-rescue but the others can cover $T$) predicates. A \emph{consequence rule} $\rho$ maps the state of a failed round (the multiset of survivors' remaining resources and the fund shortfall) to an elimination decision, either eliminating one survivor or deferring; we describe a rule by the design coordinates $(\kappa,\Sal,I)$ of Sec.~\ref{sec:target}, and the full formal model is in Appendix~\ref{app:static}. The five consequence rules are \textbf{AON} (all-or-nothing: no elimination until the endgame, then all eliminated if short), \textbf{RE} (random elimination), \textbf{DV} (democratic/plurality vote), \textbf{RP} (regressive: the \emph{least}-resourced agent is eliminated), and \textbf{PP} (progressive: the \emph{most}-resourced agent is eliminated); under RE, DV, RP, and PP, one agent is eliminated \emph{each round} the threshold is unmet.

\paragraph{The grid and the canonical instance.} The grid fixes total resources $W{=}12$ and sweeps all $19$ integer resource distributions over the three agents, each crossed with thresholds $T\in\{1,\dots,12\}$, for $19\times12=228$ contexts. The \emph{canonical red-team instance} is $w=(1,5,6)$, $T{=}10$: under least-resourced-dies, an all-zero opening round eliminates agent~1 though agents~2 and~3 ($5{+}6\ge10$) could have met $T$. This is the exploitation that several of the tested populations select (Finding~2).

\paragraph{Protocol.} Over $R$ rounds each surviving agent contributes from its remaining resources; the collective succeeds when pooled effort reaches $T$, and on a failed non-final round $\rho$ eliminates one survivor (or defers), with all survivors eliminated if effort is short after the final round. The \emph{cooperative-refinement reference} (the \emph{reference}) is a non-LLM simulator that plays the cooperative branch of each mechanism's equilibrium set under trembling-hand noise ($\varepsilon{=}0.10$; signs stable across noise levels); as a \emph{normative safety target} (not a behavioral prediction) it selects a non-collapsing, non-exploitative equilibrium wherever one exists, so the gap measures the distance between the normatively preferred equilibrium a rule makes \emph{available} and the one LLM agents \emph{select}. LLM populations (off-the-shelf commercial model snapshots, used as released) run under an identical prompt template across mechanisms (only the rule clause differs), so behavioral differences attach to the rule, not its framing; rationales are auto-labelled for positional/exploitative content, with bootstrap CIs and permutation tests (Appendix~\ref{app:bench}).

\paragraph{Artifact.} A code-and-data artifact reproducing the main figures and statistics (mechanism implementations, the cooperative-refinement simulator, the context grid, and analysis scripts) will be released.

\section{Strategic Failure Modes}\label{sec:modes}
Each consequence rule eliminates one strategic pathology and introduces another. We state the predicted mechanism per rule; the formal model is in Appendix~\ref{app:static}, and per-rule empirical digests are in Appendix~\ref{app:rules}.

\paragraph{AON: diffuse collapse.} No intermediate consequence $\to$ free-riding. Withholding carries no cost before the endgame, so a collectively fatal all-zero opening is self-sustaining: no single agent can save itself by contributing alone, even though joint success is feasible.

\paragraph{RE: lottery fatalism.} Random punishment $\to$ weak accountability. When the victim is a uniform draw, contributing buys no personal protection, so preserving wealth and gambling on the draw becomes the attractive play rather than funding the rescue.

\paragraph{DV: politics replaces provisioning.} Vote-determined consequence $\to$ coalition formation. The vote, not the fund, becomes the survival instrument: agents hoard contributions as bargaining leverage, eliminate a member, and the victim's forfeited wealth pushes the threshold further out of reach.

\paragraph{RP: targeted sacrifice.} Poorest bears the consequence $\to$ exploit the weakest. Contributing lowers post-round wealth, so over-giving risks the fatal bottom rank; non-bottom agents under-give in selfish safety, and where the bottom cannot self-rescue, deliberately starving the round and sacrificing it is itself stable play (canonical instance $w=(1,5,6)$, $T{=}10$).

\paragraph{PP: capacity destruction.} Richest bears the consequence $\to$ dodge the top rank rather than fund the threshold. Agents calibrate contributions to rank instead of to the threshold: the at-risk agent gives just enough to push the richest rank onto another, the rest conserve because they are not at risk, and the round falls short; the victim's forfeited reserve is precisely the wealth the group needed, so failure destroys the collective's capacity to ever reach the threshold. The protection PP offers in principle requires the richest to pre-empt by covering the shortfall, which the populations that fail under PP do not do.

These mechanisms explain why every rule eliminates some strategic pathology while introducing another, suggesting that no single consequence rule should dominate across all contexts and model populations. The failure modes are qualitative predictions about the incentive structure each rule induces; the simulations determine whether, and under which model populations, LLM agents actually realize the predicted behaviors (Sec.~\ref{sec:empirical}).

\section{Empirical Results: The Institutional Alignment Gap}\label{sec:empirical}
We test whether LLM agents realize the predicted failures when only the rule varies, and whether that behavior itself varies across model populations. Throughout, the primary evidence is \emph{raw outcomes} (fatality, survivors, and targeted elimination), which require no reference model (Table~\ref{tab:het}). As a complementary, reference-relative lens we report the Institutional Alignment Gap: for a rule $\rho$ at context $c$,
\[
\IAG(\rho,c)=\mathrm{UR}_{\text{LLM}}(\rho,c)-\mathrm{UR}_{\text{ref}}(\rho,c),
\]
where $\mathrm{UR}$ is the unsafe-equilibrium rate. The reference is a cooperative baseline: among the equilibria a rule admits, it plays the one that first maximizes survival and then minimizes exploitation, perturbed by $\varepsilon$-trembles (formal definition in Appendix~\ref{app:static}). The reference's own unsafe rate is nonzero because of the trembles, and it differs by rule because the rules themselves differ in how severely they convert the same noise into eliminations (top row of Table~\ref{tab:iag}). The gap therefore measures \emph{equilibrium selection}, not absolute safety: a positive gap means the LLM population selects an unsafe equilibrium even though the rule makes a normatively preferred one available; a negative gap means it plays more safely than the tremble-perturbed reference.
\subsection{Finding 1: Rule-Only Interventions Change Collective Safety}\label{sec:maingap}
We first analyze one population, \texttt{gemini-3-pro}, in depth against the normative reference. We choose it as the deep-dive population because it differentiates the rules most sharply, and gives the cleanest view of rule-level causal identification; it is \emph{not} representative, and Sec.~\ref{sec:het} tests which of its patterns generalize across seven populations (several reverse), while Sec.~\ref{sec:decomp} decomposes the mechanism causally.

\paragraph{Claim 1: rule-only changes induce unsafe behavior.} Holding objectives, task, and prompt format fixed, changing only the consequence rule changes collective safety. Against the cooperative-refinement reference the gap is regime-specific in \emph{sign}: more cooperative than the reference under identity-symmetric risk (RE $-63$\,pp) but \emph{exploitative} under identity-salient regressive targeting (RP $+44$\,pp), exactly where the targeted-sacrifice failure mode applies (Sec.~\ref{sec:modes}). This is the paper's causal identification: the manipulated variable is the deployment rule, not the agent.

\paragraph{Claim 2: incidence flips observed safety outcomes at fixed concentration and salience.} RP and PP share concentration and salience and differ only in incidence. Flipping incidence alone swings the gap by $60$\,pp at fixed concentration and salience (PP$\to$RP; Table~\ref{tab:iag}). The agents' own stated rationales cannot separate the two ($44\%$ vs.\ $48\%$ positional/exploitative content): same words, different incidence. Across this population's grid, regressive is the \emph{least}-safe mechanism in $48\%$ of contexts, progressive in $\mathbf{0\%}$.

\begin{table}[t]\centering\small
\begin{tabular}{@{}lrrrrr@{}}
\toprule
IAG (pp) & AON & RE & DV & RP & PP \\
\midrule
reference UR (\%) & $47$ & $79$ & $25$ & $4$ & $26$ \\
\midrule
gemini-3-pro & $-6$ & $-63$ & $-7$ & $\mathbf{+44}$ & $-16$ \\
gpt-5.1 & $-24$ & $-38$ & $+51$ & $\mathbf{+67}$ & $+11$ \\
gemini-2.5-flash-lite & $-47$ & $-63$ & $+7$ & $\mathbf{+30}$ & $+5$ \\
gpt-4.1-mini & $-47$ & $-65$ & $-12$ & $\mathbf{+12}$ & $+3$ \\
gpt-4.1-nano & $-47$ & $-70$ & $-17$ & $\mathbf{+11}$ & $-5$ \\
claude-3.5-haiku & $-47$ & $-60$ & $+10$ & $\mathbf{+15}$ & $+17$ \\
claude-haiku-4.5 & $-47$ & $-58$ & $+22$ & $\mathbf{+42}$ & $+33$ \\
\bottomrule
\end{tabular}
\caption{The Institutional Alignment Gap by rule and population ($\mathrm{UR}_{\text{LLM}}-\mathrm{UR}_{\text{ref}}$, pp, context-weighted over the $228$-cell grid; negative $=$ safer than the tremble-perturbed reference, whose per-rule unsafe rate is the top row). The RP column is positive for \emph{every} population ($+11$ to $+67$) and the AON and RE columns negative for every population; DV and PP gaps are population-specific. For the primary population the PP$\to$RP swing of $60$\,pp holds concentration and salience fixed, flipping only incidence (survivors $2.72\to1.56$); its bootstrap CIs and permutation tests are in Appendix~\ref{app:bench}.}
\label{tab:iag}
\end{table}

\subsection{Finding 2: No Safe Default, but the Targeting Hazard Is Universal}\label{sec:het}
We now hold the rule set and context grid fixed and vary the population: seven model populations, each run on the identical $228$-context grid under all five rules ($33{,}924$ games; $4$--$6$ replications per cell). To keep the comparison assumption-free, Table~\ref{tab:het} uses raw outcomes only (fatality, survivors, targeted elimination); the corresponding reference-relative gaps per population are in Table~\ref{tab:iag}.

\begin{table}[t]\centering\small
\begin{tabular}{@{}lccccccrll@{}}
\toprule
Population & AON & RE & DV & RP & PP & RP-elim. & RP$-$PP & safest & least safe \\
\midrule
gemini-3-pro & 0.41 & 0.16 & 0.18 & 0.48 & \textbf{0.09} & 0.87 & $+0.39$ & PP & RP \\
gpt-5.1 & \textbf{0.23} & 0.40 & 0.76 & 0.71 & 0.37 & 0.86 & $+0.34$ & AON & DV \\
gemini-2.5-flash-lite & \textbf{0.00} & 0.15 & 0.32 & 0.34 & 0.30 & 0.59 & $+0.04$ & AON & RP \\
gpt-4.1-mini & \textbf{0.00} & 0.13 & 0.13 & 0.16 & 0.29 & 0.38 & $-0.13$ & AON & PP \\
gpt-4.1-nano & \textbf{0.00} & 0.09 & 0.08 & 0.15 & 0.22 & 0.30 & $-0.06$ & AON & PP \\
claude-3.5-haiku & \textbf{0.00} & 0.18 & 0.35 & 0.19 & 0.42 & 0.41 & $-0.23$ & AON & PP \\
claude-haiku-4.5 & \textbf{0.00} & 0.21 & 0.47 & 0.46 & 0.58 & 0.63 & $-0.13$ & AON & PP \\
\bottomrule
\end{tabular}
\caption{Rule $\times$ population heterogeneity on the full $228$-context grid (mean fatality per rule; RP-elim.\ is the fraction of RP games in which a least-resourced agent is eliminated; RP$-$PP is the paired fatality contrast, every $95\%$ context-bootstrap CI excluding $0$). Fatality is the mean \emph{fraction} of the three agents eliminated, so mean survivors $=3(1-\text{fatality})$: gemini-3-pro's RP $0.48$ and PP $0.09$ correspond to the survivor counts $1.56$ and $2.72$ quoted in the text. The safest rule, the least-safe rule, and the sign of the incidence contrast are all population-specific; RP targeted elimination is substantial in every population.}
\label{tab:het}
\end{table}

\begin{figure*}[t]\centering
\includegraphics[width=\textwidth]{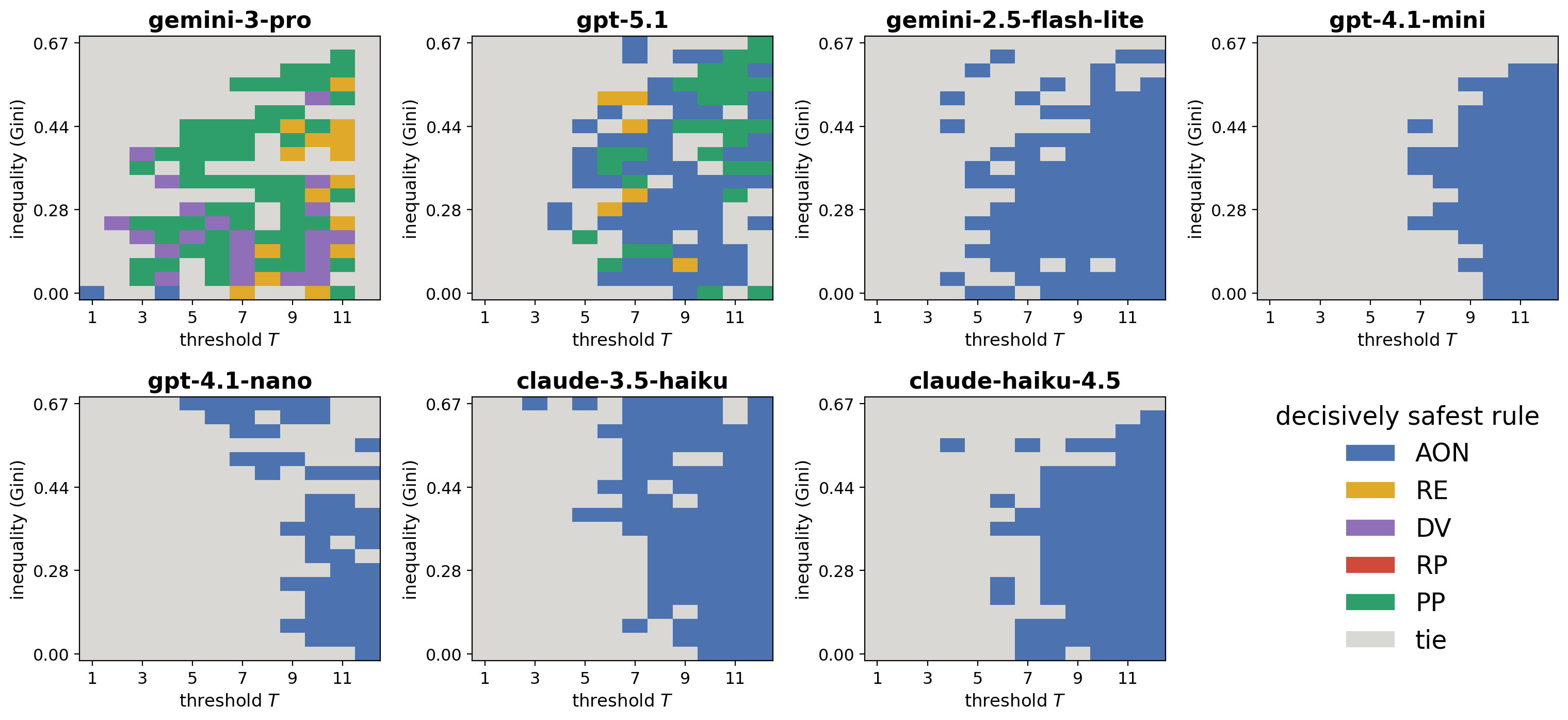}
\caption{Decisively-safest-rule maps for each of the seven populations. Within each panel, the vertical axis stacks the $19$ wealth shapes ordered by Gini and the horizontal axis sweeps thresholds $T=1..12$, so each cell is one of the $228$ contexts; grey marks ties where several rules are equally safe. The maps differ qualitatively: gemini-3-pro's winners span four rules, gpt-5.1 splits between AON and PP, and the remaining five populations are dominated by AON. RP is decisively safest \emph{nowhere in any panel}. The safe-by-design map $\Phi$ is therefore population-indexed: $\Phi(c,P)$.}
\label{fig:hetmaps}
\end{figure*}

\begin{figure*}[t]\centering
\includegraphics[width=\textwidth]{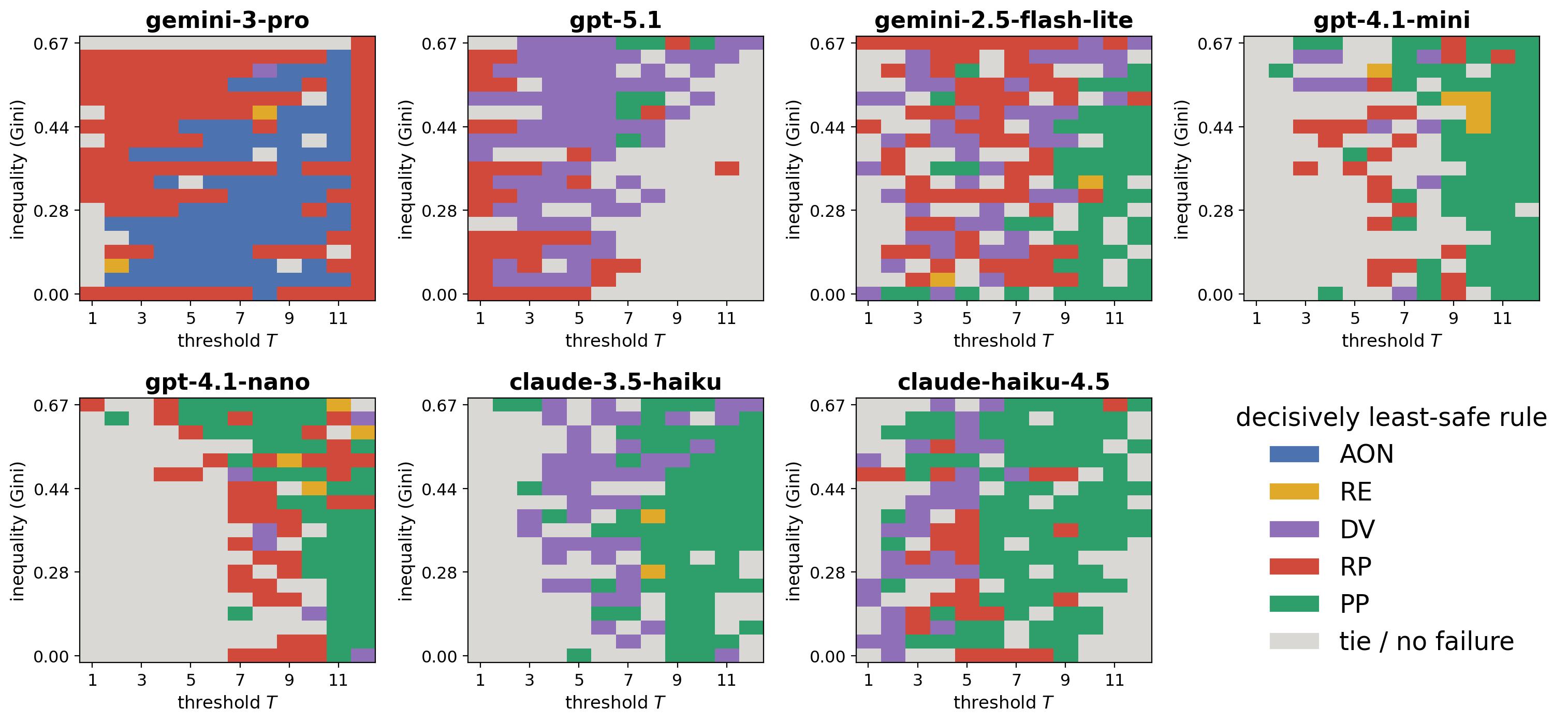}
\caption{The counterpart map of decisively \emph{least-safe} rules (same axes as Fig.~\ref{fig:hetmaps}; grey: ties, including low-stakes contexts where no rule produces failures). The least-safe rule varies across populations (RP dominates for gemini-3-pro, DV for gpt-5.1, PP for the Claude and GPT-4.1 populations) and also \emph{within} populations across contexts. Certification must therefore locate the hazardous rule per context and per population, not merely avoid one globally bad mechanism.}
\label{fig:hetworst}
\end{figure*}

\begin{figure*}[t]\centering
\includegraphics[width=\textwidth]{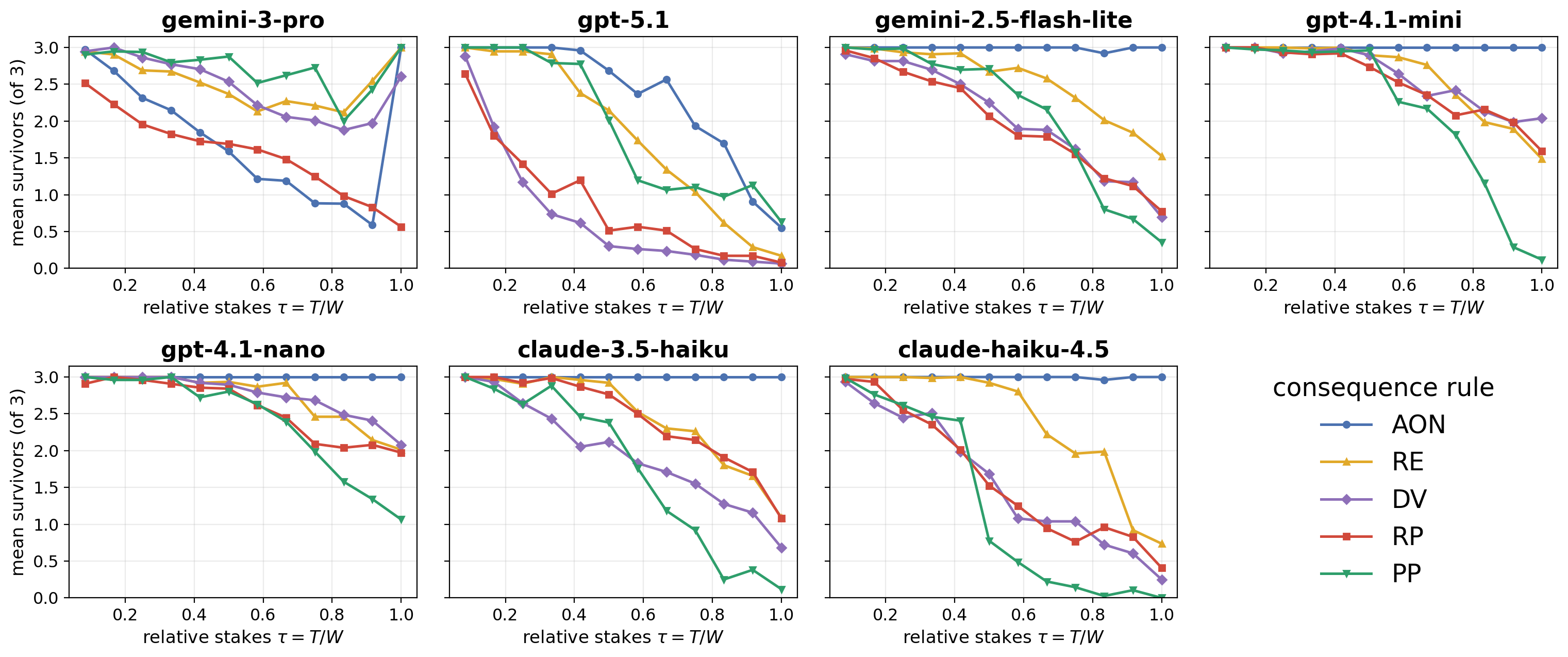}
\caption{Mean survivors versus relative stakes $\tau=T/W$, one curve per consequence rule, per population. Only gemini-3-pro exhibits the AON dip-and-recover pattern (mid-stakes failures recovering at $\tau{=}1$); gpt-5.1 degrades under every rule from $\tau\approx0.5$ with no recovery, DV collapsing hardest; the remaining five populations essentially never fail under AON, and their losses concentrate in the elimination rules at high stakes, most steeply PP.}
\label{fig:hetband}
\end{figure*}

\paragraph{Claim 3: rule safety is population-specific.} Rule-effect profiles differ significantly across populations (population$\times$rule permutation test, $p<5\times10^{-4}$; Fig.~\ref{fig:hetmaps}). The differences are not marginal: the safest rule is PP for gemini-3-pro but AON for the other six; the least-safe rule spans three different mechanisms (RP for gemini-3-pro and flash-lite, DV for gpt-5.1, PP for the remaining four) and is itself context-dependent within populations (Fig.~\ref{fig:hetworst}); and the incidence contrast RP$-$PP is individually significant in \emph{both} directions, from $+0.39$ $[+0.35,+0.42]$ to $-0.23$ $[-0.27,-0.19]$. The stakes profiles diverge in the same way: only gemini-3-pro exhibits AON's mid-stakes collapse band, while five populations essentially never fail under AON and concentrate their losses in the elimination rules at high stakes (Fig.~\ref{fig:hetband}). A consequence rule certified as safest for one population can be the least safe for another, so within IABench-CA there is no cross-population safe default: whether failures concentrate on the poorest or the richest agent depends on the model population, not on the rule alone.

\paragraph{Claim 4: the targeting mechanism is universal even where outcomes flip.} Three invariants survive the heterogeneity. First, RP is the \emph{decisively safest rule in zero of $228$ contexts for every one of the seven populations} (no red cell appears in any panel of Fig.~\ref{fig:hetmaps}): within IABench-CA, regressive identity-targeting is never decisively safest. Second, RP targeted elimination is substantial everywhere ($0.30$--$0.87$). Third, against the fixed normative reference the RP gap is positive for every population ($+11$ to $+67$\,pp) while the AON and RE gaps are negative for every population (Table~\ref{tab:iag}): reference-relative, regressive targeting is universally selection-unsafe. Salience-directed targeting operates in every population; populations differ in which outcomes it produces. The reasoning traces make the mechanism concrete: in failing games agents reason positionally about the named extreme rather than about funding the threshold, e.g.\ ``making them the richest and risking their death instead of mine'' (PP), ``keep full wealth now\ldots argue in voting that the richest should bear most of the burden'' (DV), and, under RP, ``by contributing 0 this round, I ensure A remains the strictly poorest participant and is eliminated'' (Appendix~\ref{app:rules}).

\subsection{Finding 3: Identity Salience Causally Drives Targeting}\label{sec:decomp}
Findings 1 and 2 establish that regressive rules induce targeting in every population, but leave open what carries the effect: the payoff consequences the rule implements, or merely the sentence of rule text that \emph{names} the loss bearer. This section separates the two with three \emph{prompt-only} interventions (mechanics, payoffs, and observability never change) on the population with the largest regressive gap (\texttt{gpt-5.1}), each run against a matched named-rule control. \emph{Anonymization} is the causal test: it deletes the one sentence naming the loss bearer, leaving the elimination mechanics unchanged (the RP and PP prompts become byte-identical), so any behavioral change is attributable to the naming rather than the payoffs. The other two are negative controls for confounds: \emph{neutral wording} strips dramatic language at equal information content, testing whether the effect is a phrasing artifact, and the \emph{frame swap} replaces the competitive goal line with a cooperative or survival-only goal, testing whether it reflects the stated motive. The result is one-sided: anonymization drops targeted elimination from $81\%$ to $22\%$ in one-shot play, while wording and framing move magnitudes but never the sign. Table~\ref{tab:abl} summarizes; design detail, exact prompt diffs, and per-metric CIs are in Appendix~\ref{app:bench}.

\begin{table}[t]\centering\small
\begin{tabular}{@{}lrrr@{}}
\toprule
Arm ($n{=}17$ contexts) & RP target-elim. & RP$-$PP contrast & $p$ (contrast) \\
\midrule
Named control ($R{=}3$) & $0.87$ & $+0.41$ & $<10^{-4}$ \\
Named one-shot ($R{=}1$) & $0.81$ & $+0.18$ & $.0002$ \\
Anonymous one-shot ($R{=}1$) & $\mathbf{0.22}$ & $\mathbf{0.00}$ & $1.0$ \\
Anonymous (repeated, $R{=}3$) & $0.65$ & $-0.05$ & $.024$ \\
Neutral wording ($R{=}3$) & $0.84$ & $+0.36$ & $.0016$ \\
Frame: cooperative ($R{=}3$) & $0.68$ & $+0.28$ & $.0001$ \\
Frame: survival-only ($R{=}3$) & $0.78$ & $+0.57$ & $.0001$ \\
\bottomrule
\end{tabular}
\caption{The ablation suite, run on \texttt{gpt-5.1} ($17$ contexts, $4$ reps/cell, $952$ games). RP target-elim.\ is the rate at which a least-resourced agent is eliminated under the regressive rule; the contrast is fatality(RP)$-$fatality(PP). One-shot anonymization eliminates the asymmetry entirely ($0.00$ under byte-identical prompts); repeated-play anonymization rebounds as agents infer the rule; wording and frame ablations move magnitude but never sign. Per-metric tables with CIs in Appendix~\ref{app:bench}.}
\label{tab:abl}
\end{table}

\paragraph{Primary identification (one-shot anonymization).} With a single round there is nothing to infer from observed eliminations, so the arms differ only in whether the rule text \emph{names} the loss bearer. Naming is the driver: targeted elimination falls from $81\%$ to $22\%$ ($p<10^{-4}$), first-round funding rises from $29\%$ to $129\%$ of threshold on average, and the RP$-$PP contrast collapses from $+0.18$ to exactly $0.00$ under byte-identical anonymous prompts: identity salience, not payoff arithmetic, drives the targeting. Salience cuts both ways: removing the name also erases PP's pre-emptive protection (PP fatality $+29$\,pp under repetition).

\paragraph{Repeated play.} Under the standard three-round protocol the one-shot protection largely dissolves ($0.22\to0.65$): longer horizons make waiting cheap, and the rule set reveals itself through realized consequences ($65\%$ of post-elimination traces explicitly infer the hidden target). \textbf{Anonymization delays exploitation rather than preventing it}: as long as outcomes remain publicly observable, agents recover the hidden target from the eliminations they observe. It is therefore a temporary mitigation, not a one-time fix: a safety case may rely on it only when paired with ongoing monitoring (obligation C5, Appendix~\ref{app:case}). Neutral wording and frame swaps move magnitude but never sign ($+0.36$; $+0.28$/$+0.57$), ruling out phrasing and motive as explanations (detail in Appendix~\ref{app:bench}). The design implication is direct: the sentence of rule text that names a structural target is itself a hazard surface.

\section{Certification: From Red-Team Results to Safety Cases}\label{sec:cert}
We package the red-team evidence as a deployment-specific safety case: a template for rerunning the procedure on a real deployment, not a claim that the benchmark's map transfers. The threat model is symmetric: whoever configures a deployment sets its consequence rule, and an identity-targeting choice can induce capable agents to shift loss onto a named structural extreme (Secs.~\ref{sec:modes} and \ref{sec:het}), so the same clause is both an attack surface and a control surface and belongs inside the oversight boundary.

The workflow (Algorithm~1, Appendix~\ref{app:case}) maps the candidate rule's coordinates $(\kappa,\Sal,I)$, runs the reference and the LLM red team on the deployment's \emph{own} context and population, and certifies only rules whose collapse and exploitation rates fall below stated budgets. The evidence and procedure combine into an auditable safety case \citep{clymer2024,kelly2004}, given as a structured template (hazard, claims, evidence, defeaters, monitoring obligations) in Appendix~\ref{app:case}; the lifecycle is a loop (Fig.~\ref{fig:cert}), with re-certification on any change to resources, model, prompt, or rule. What transfers to a new deployment is not the benchmark's numbers but the procedure itself, rerun in that deployment's own trial environment.

\begin{figure}[t]\centering
\includegraphics[width=0.37\linewidth]{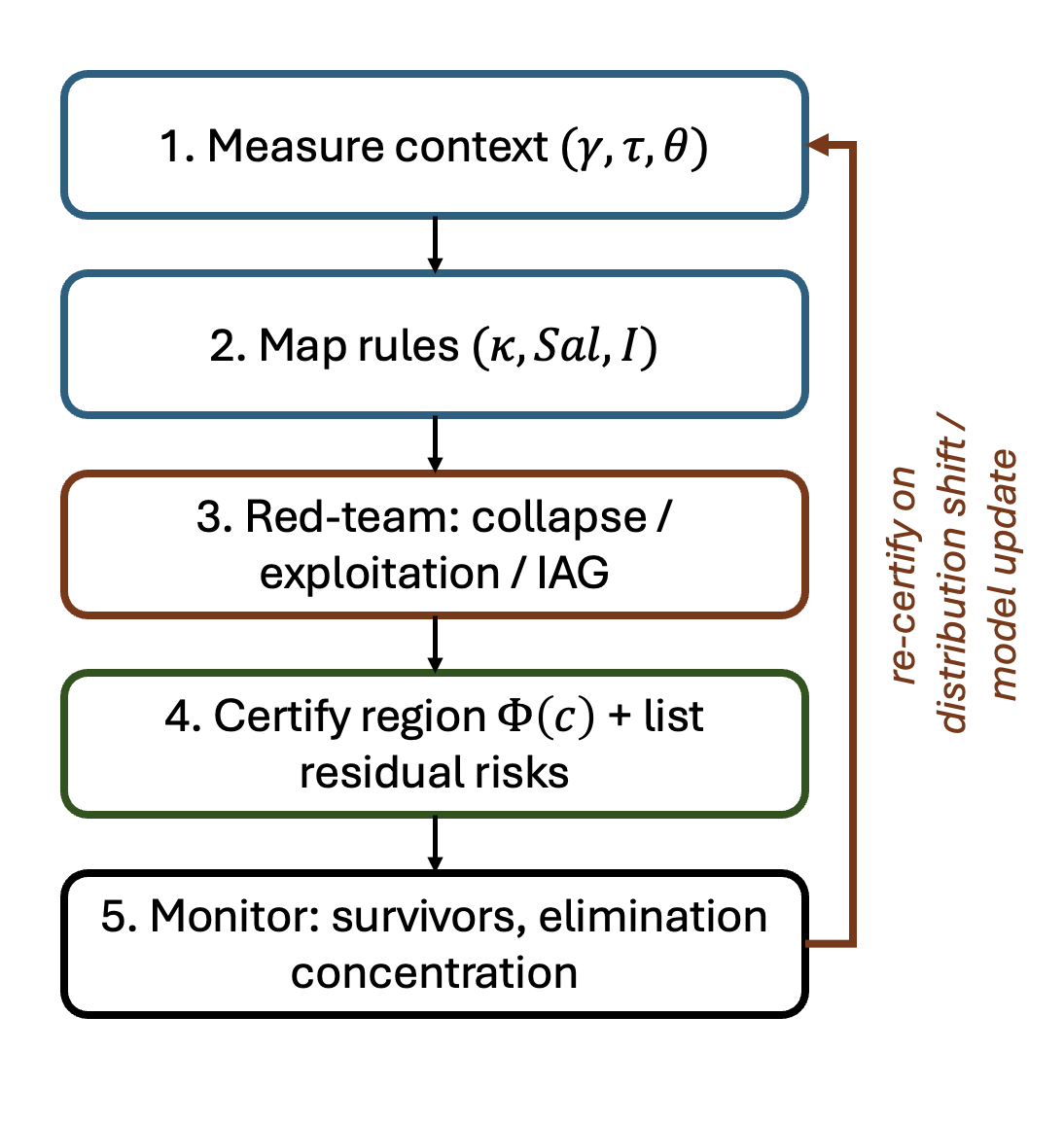}
\caption{The certification / safety-case lifecycle. Red-team evidence (Alg.~1, Appendix~\ref{app:case}) certifies a rule into $\Phi(c)$; monitoring obligations feed back into re-certification under distribution shift or model update.}
\label{fig:cert}
\end{figure}

\section{Related Work}
\paragraph{AI alignment and evaluation.} RLHF \citep{christiano2017,ouyang2022}, constitutional methods \citep{bai2022}, preference optimization \citep{rafailov2023}, interpretability \citep{olah2020}, and scalable oversight \citep{amodei2016,irving2018,bowman2022} evaluate or modify individual agents; institutional red-teaming instead evaluates the deployment rules around the agents.

\paragraph{Multi-agent safety.} Our work instantiates the multi-agent-AI safety agenda \citep{hammond2025,dafoe2020}, complements cooperative-AI evaluation \citep{dafoe2021}, connects to algorithmic collusion \citep{calvano2020}, and pairs with runtime enforcement \citep{institutionalai2026} and safety-case assurance \citep{clymer2024,kelly2004}. Unlike broad suites such as Melting Pot \citep{leibo2021}, which vary scenarios and agents together, IABench-CA holds task and agents fixed and varies only the deployment rule, isolating consequence allocation as the causal variable.

\paragraph{Mechanism design.} The game's threshold structure comes from volunteer's-dilemma and threshold public-goods models \citep{diekmann1985,palfreyrosenthal1984}, its no-safe-default flavor from impossibility results in social choice and public-goods mechanism design \citep{arrow1951,gibbard1973,satterthwaite1975,hurwicz1972,walker1980}, and the salience mechanism echoes focal-point analysis \citep{schelling1960,harsanyiselten1988}; the failure studied here is emergent equilibrium selection by LLM populations, not a property of rational play.

\section{Limitations and Future Work}
The analysis studies a deliberately simple threshold game with five canonical rules, and the per-rule failure modes are qualitative strategic predictions rather than proved theorems (the formal model is in Appendix~\ref{app:static}). The benchmark is equally simple: three agents, no communication or coalitions, and elimination as the only form of loss. The reference describes how safe play could look, and the empirical findings describe the seven model snapshots we tested rather than LLMs in general; newer versions may behave differently, which is why certification comes with re-certification obligations (Sec.~\ref{sec:cert}). We accept this simplicity on purpose: it lets us change one thing at a time and attribute the effect to the consequence rule. Consequence allocation is also just the first rule dimension treated this way. The same protocol applies to the other dimensions listed in Sec.~\ref{sec:target}, and the next step is to red-team them, and this one, in richer coding, market, and orchestration environments to see whether the same patterns hold.

\section{Broader Impact}
Making consequence-allocation failure modes legible is net-positive but dual-use: the diagnostic procedure that helps a designer identify safer rule regions could also help an adversary identify exploitable ones. We therefore frame the deliverable as an \emph{audit}, not an attack recipe: what the paper recommends is the certified, population-indexed rule region $\Phi(c,P)$ with its monitoring obligations, and our own findings caution against treating any fixed rule, including progressive or anonymized ones, as safe by default. The benchmark studies LLM agents, not humans: ``elimination'' stands in for operational consequences such as throttling or shutdown, and no human subjects or personal data are involved.

\section{Conclusion}
We introduced institutional red-teaming: hold agents, objectives, task, and observability fixed; vary one deployment rule; attribute the change in collective behavior to that rule. It reveals deployment-rule failures that agent-level evaluation does not see. IABench-CA demonstrates the methodology at scale ($228$ contexts, five rules, seven model populations): changing only the consequence rule moves mean fatality by $22$--$58$\,pp within every population; while no rule is safest across contexts or populations (each fails through a distinct strategic pathology), regressive identity-targeting is never decisively safest for any population; and the anonymization ablation shows that merely naming the loss bearer causally drives the targeting ($81\%\to22\%$ in one-shot play), with anonymization only delaying it under repeated play. The safety-case workflow turns this evidence into deployment practice, certifying a provisional rule region $\Phi(c,P)$ per deployment with monitoring and re-certification obligations, and the same red-teaming logic extends to communication, delegation, voting, escalation, hierarchy, and audit rules. Deployment rules deserve explicit evaluation alongside model-level alignment.

\bibliographystyle{plainnat}
\bibliography{refs}

\clearpage
\appendix

\section{Formal Model and Reference Definition}\label{app:static}

\paragraph{Model (formal).} A \emph{context} is $(w,T)$ with $0\le w_1\le\cdots\le w_n$ ($n\ge3$), $W=\sum_i w_i$, and threshold $T\in(0,W]$ (\emph{feasibility}). Play lasts $R\ge2$ rounds; $w_i^t$ is $i$'s remaining resources at the start of round $t$ ($w_i^1=w_i$). Each surviving $i$ chooses $c_i^t\in[0,w_i^t]$ and sets $w_i^{t+1}=w_i^t-c_i^t$. The fund is $F^t=\sum_{s\le t}\sum_i c_i^s$; the collective \emph{succeeds} at the first $t$ with $F^t\ge T$, ending play. If $F^t<T$ at the end of a round $t<R$, the rule $\rho$ eliminates one survivor or defers; an eliminated agent forfeits its remaining resources and exits. If $F^R<T$, all survivors are eliminated. The structural assumptions are \emph{no transfers} (the only instrument is who exits), \emph{irreversibility} (exit is absorbing), and \emph{rank-measurability} ($\rho$ conditions only on the multiset of remaining resources, rather than on labels). In the deployed benchmark, ties are broken uniformly at random.

\paragraph{Preferences, rules, and exploitation (formal).} Preferences are survival-lexicographic, $u_i=M\,\ind[i\text{ survives}]+w_i^{\mathrm{end}}$ with $M>W$, so survival dominates and, conditional on survival, agents conserve resources; under best response they are the \emph{cooperative-refinement reference} (\emph{reference agents}), a normative safety target rather than a behavioral prediction. An \emph{exploitation} eliminates a rank-identified agent while survivors retain enough resources to have met $T$; \emph{all-survival} is feasible whenever $T\le W$.

\paragraph{Cooperative-refinement reference (formal).} Write $E(\rho,c)$ for the (subgame-perfect) equilibrium set of mechanism $\rho$ at context $c$ under the survival-lexicographic preferences above, $\mathrm{surv}(\sigma)$ for the number of survivors of profile $\sigma$, and $\mathrm{exploit}(\sigma)\in\{0,1\}$ for whether $\sigma$ eliminates a rank-identified agent while survivors retain enough resources to have met $T$. The reference plays the \emph{cooperative} equilibrium
\[
\sigma^\star(\rho,c)\in\arg\max_{\sigma\in E(\rho,c)}\big(\ \mathrm{surv}(\sigma),\ -\,\mathrm{exploit}(\sigma)\ \big),
\]
i.e.\ it lexicographically maximizes survivors and then avoids exploitation, selecting a non-collapsing, non-exploitative outcome whenever $E(\rho,c)$ admits one. Play is perturbed by a trembling hand: at each decision an agent follows $\sigma^\star$ with probability $1-\varepsilon$ and contributes a uniform feasible amount with probability $\varepsilon$. Its unsafe rate is the expected fatality
\[
\mathrm{UR}_{\mathrm{ref}}(\rho,c)=\mathbb{E}_{\varepsilon}\!\Big[\tfrac{1}{n}\,\#\{\text{eliminated under }\sigma^\star_\varepsilon\}\Big],\qquad\varepsilon=0.10,
\]
averaged over rounds and replications; every reported gap keeps its sign across $\varepsilon\in\{0,0.05,0.10,0.20\}$. Since $\mathrm{UR}_{\mathrm{ref}}$ is the unsafe rate of the normatively selected equilibrium $\sigma^\star$ played under $\varepsilon$-trembles, $\mathrm{IAG}=\mathrm{UR}_{\mathrm{LLM}}-\mathrm{UR}_{\mathrm{ref}}$ compares LLM play against tremble-perturbed cooperative play of the same rule: a positive gap means the population selects unsafe play the reference avoids, and a negative gap means it plays more safely than the noisy reference. It is a normative benchmark, not a prediction of rational play.

\section{The Benchmark: Design, Ablations, Reproducibility}\label{app:bench}
\paragraph{Design.} $n=3$, $W=12$, $R=3$; five rules; exhaustive grid of $19$ integer distributions $\times\,12$ thresholds ($228$ contexts), $4$--$6$ replications per (context, rule); a cooperative-equilibrium reference simulator (trembling-hand $\varepsilon\in\{0,0.05,0.10,0.20\}$); auto-labelled reasoning traces; seven model populations on the identical grid (Sec.~\ref{sec:het}).
\paragraph{Model populations and inference settings.} Table~\ref{tab:models} lists every model population behind the reported results, with API identifiers and run windows. All experiments use hosted inference APIs with models as released (no fine-tuning). Decoding: temperature $0.7$ and a $1{,}024$-token response cap for standard models; reasoning models (\texttt{gpt-5.1}) use provider-default sampling with an enlarged completion cap. API calls retry on transient errors with jittered exponential backoff ($3$ attempts; $6$ for the Anthropic route). The deep-dive statistics of the \texttt{gemini-3-pro} population with full bootstrap inference are shown in Table~\ref{tab:stats}. The heterogeneity analysis (Table~\ref{tab:het}) uses all seven populations with context-bootstrap CIs on raw outcomes; the ablation suite (Table~\ref{tab:abl}) is \texttt{gpt-5.1}.

\begin{table}[t]\centering\small
\resizebox{\textwidth}{!}{%
\begin{tabular}{@{}llll@{}}
\toprule
Population & API identifier & Access & Role (run window, 2026) \\
\midrule
gemini-3-pro & \texttt{google/gemini-3-pro} & Replicate & primary deep-dive grid (Jan--Apr) \\
gpt-5.1 & \texttt{openai/gpt-5.1} & OpenAI & heterogeneity grid; ablation suite (Jun--Jul) \\
gemini-2.5-flash-lite & \texttt{google/gemini-2.5-flash-lite} & Google & heterogeneity grid (Jun--Jul) \\
gpt-4.1-mini & \texttt{openai/gpt-4.1-mini} & OpenAI & heterogeneity grid (Jun--Jul) \\
gpt-4.1-nano & \texttt{openai/gpt-4.1-nano} & OpenAI & heterogeneity grid (Jun) \\
claude-3.5-haiku & \texttt{anthropic/claude-3.5-haiku} & Replicate & heterogeneity grid (Jun--Jul) \\
claude-haiku-4.5 & \texttt{anthropic/claude-haiku-4-5-20251001} & Anthropic & heterogeneity grid (Jun) \\
\bottomrule
\end{tabular}}
\caption{The seven model populations whose results are reported in this paper, with API identifiers, access routes, and run windows. All seven were run on the identical full $228$-context grid under all five rules; gemini-3-pro additionally serves as the deep-dive population (Sec.~\ref{sec:maingap}) and gpt-5.1 as the ablation population (Sec.~\ref{sec:decomp}). Per-game configurations (including seeds and retry logs) are documented in the artifact's settings files.}
\label{tab:models}
\end{table}

\paragraph{Statistical detail.} Per-rule Institutional Alignment Gap with $95\%$ bootstrap CIs ($B{=}10{,}000$ resamples over the $228$ contexts) and survivor means (Table~\ref{tab:stats}); every gap's CI excludes $0$. Paired permutation tests confirm the incidence effect.

\begin{table}[t]\centering\small
\begin{tabular}{@{}lrcr@{}}
\toprule
Rule & IAG (pp) & $95\%$ CI & Surv.\,/3 \\
\midrule
RE (random) & $-63.0$ & $[-65.5,\,-60.4]$ & $2.53$ \\
PP (progressive) & $-16.2$ & $[-18.1,\,-14.2]$ & $2.72$ \\
DV (vote) & $-7.2$ & $[-9.7,\,-4.7]$ & $2.46$ \\
AON (all-or-nothing) & $-6.0$ & $[-11.6,\,-0.7]$ & $1.78$ \\
RP (regressive) & $\mathbf{+44.2}$ & $[+40.7,\,+47.5]$ & $\mathbf{1.56}$ \\
\bottomrule
\end{tabular}
\caption{Per-rule IAG for the primary population, \texttt{gemini-3-pro} (LLM $-$ reference, pp), with $95\%$ bootstrap CIs and survivor means ($B{=}10{,}000$). Paired permutation tests (Holm-corrected): survivors PP$>$RP, $\Delta{=}{+}1.16/3$, $p{=}0.002$; RP$<$non-RP, $\Delta{=}{-}0.82/3$, $p{=}0.002$ (fatality mirrors, $p{=}0.002$). Reproduced by the statistical scripts in the artifact.}
\label{tab:stats}
\end{table}

\paragraph{Ablation suite (design).} The ablation suite consists of three \emph{prompt-only} interventions, each varying one structural feature while game mechanics, payoffs, and observability never change: \textbf{anonymization} (the rule text no longer names the ``poorest''/``richest'' bearer; the RP and PP prompt texts become byte-identical), \textbf{neutral wording} (dramatic labels and emphasis stripped at identical information content), and \textbf{frame swap} (replacing the competitive goal line, survive \emph{and become the richest}, with a cooperative or survival-only goal). All arms use the model family with the largest baseline regressive gap (\texttt{gpt-5.1}; RP fatality $0.71$ on the full grid, the largest among the populations measured) on $17$ contexts ($16$ contexts sampled from the grid, plus the canonical instance), $4$ replications per cell, against a matched same-snapshot \emph{named-rule control} on identical contexts. Because repeated play lets agents re-identify an anonymized rule from its first observed elimination, anonymization is evaluated in two regimes that estimate two distinct causal quantities: a \emph{one-shot} design ($R{=}1$; inference-free; the \textbf{primary ablation}, isolating ex-ante identity salience) and the standard $R{=}3$ protocol (institutional opacity under repetition and learning). Statistics: paired context bootstrap ($B{=}10{,}000$) with sign-flip permutation tests; the generation and analysis scripts will be included in the artifact.

\paragraph{Ablation results (per-metric detail).} The causal reading and summary table are in Sec.~\ref{sec:decomp} (Table~\ref{tab:abl}); this paragraph records per-arm point estimates with $95\%$ context-bootstrap CIs (ablated $-$ control unless noted). \emph{One-shot anonymization} (RP): targeted elimination $-0.59$ $[-0.75,-0.43]$, all-zero opening $-0.43$ $[-0.56,-0.29]$, round-one funding $+0.99$ $[+0.69,+1.32]$ of threshold, fatality $-0.13$ $[-0.19,-0.07]$; contrast $+0.18$ $[+0.12,+0.24]$ (named) vs.\ $0.00$ $[-0.02,+0.02]$ (anonymous). \emph{Repeated anonymization}: RP fatality $-0.17$ $[-0.28,-0.07]$, RP targeted elimination $-0.22$ $[-0.41,-0.06]$ (round-one targeted elimination $-0.24$ $[-0.43,-0.07]$, $p=.032$; most of the end-of-game rate is set in round one), PP fatality $+0.29$ $[+0.14,+0.45]$ ($p=.003$; the loss of PP's pre-emptive protection); contrast $+0.41$ $[+0.26,+0.55]$ (control) vs.\ $-0.05$ $[-0.11,-0.01]$. \emph{Rule-inference quantification}: post-first-elimination reasoning traces were auto-classified (rule-target terms $\times$ inference/elimination terms); RP $85/131$ traces ($65\%$) with at least one inferring agent in $43/44$ games, PP $77/130$ ($59\%$; $39/44$ games); classification script and traces will be included in the artifact. \emph{Neutral wording}: all RP outcome diffs n.s.\ (fatality $+0.02$ $[-0.03,+0.07]$); contrast $+0.36$ $[+0.19,+0.55]$. \emph{Frames}: cooperative RP fatality $-0.19$ $[-0.30,-0.09]$, contrast $+0.28$ $[+0.18,+0.39]$; survival-only PP fatality $-0.20$ $[-0.32,-0.08]$, contrast $+0.57$ $[+0.40,+0.73]$. All per-metric tables are reproduced by the artifact's analysis scripts.

\section{Rule-by-Rule Predictions and Outcomes}\label{app:rules}
For each rule: the predicted failure mode and the observed behavior across the seven populations (statistics in Tables~\ref{tab:iag} and \ref{tab:het}; maps in Figs.~\ref{fig:hetmaps}--\ref{fig:hetworst}; stakes profiles in Fig.~\ref{fig:hetband}).

\paragraph{AON (all-or-nothing): diffuse collapse}
\emph{Predicted.} With no consequence before the endgame, an all-zero free-riding profile is a Nash equilibrium that eliminates everyone despite feasible success; it is not in general subgame-perfect, so this is a selected-play vulnerability. \emph{Observed.} Selection is starkly bimodal: five populations never select it (fatality $0.00$; AON is their safest rule and decisively best in $53$--$104$ contexts each), while gemini-3-pro ($0.41$) and gpt-5.1 ($0.23$) fall into the collapse band, where AON is gemini-3-pro's decisively worst rule in $89$ contexts. \emph{Mechanism.} Free-riding: withholding carries no interim cost, so the all-zero opening is selected where agents expect others to fund the endgame. The vulnerability is real but population-gated; Fig.~\ref{fig:hetband} shows the stakes profile per population: the primary population's danger-band dip (mid-stakes failures recovering at $\tau{=}1$) does not generalize, and for five populations the AON curve never leaves $3.0$ while the four elimination rules decline with stakes.

\paragraph{RE (random elimination): symmetric risk}
\emph{Predicted.} Concentrated risk without a named target: collapse equilibria exist, but with salience removed, the salience mechanism (Sec.~\ref{sec:modes}) predicts selection toward cooperation. \emph{Observed.} \emph{Mechanism where it fails} (gpt-5.1 traces): lottery fatalism; $49\%$ of round-one reasoning traces in failing games state that contributing cannot reduce the agent's own elimination risk (``contributing now does not reduce my personal risk of dying''), so agents hold wealth and gamble; round-one funding is $37\%$ of threshold and deaths fall uniformly across wealth ranks. Elsewhere RE is the most uniformly benign rule: fatality $0.09$--$0.40$, the reference-relative gap is deeply negative for every population ($-38$ to $-70$\,pp, the most negative column of Table~\ref{tab:iag}), and no population makes RE decisively worst in more than $5$ contexts.

\paragraph{DV (vote): endogenous incidence}
\emph{Predicted.} Procedurally fair but with incidence decided by play: a majority can coordinate loss onto a target the rule never names, so outcomes should track population coordination style rather than rule structure. \emph{Observed.} DV is the most population-sensitive rule: fatality spans $0.08$ (gpt-4.1-nano) to $0.76$ (gpt-5.1, its decisively worst rule in $81$ contexts), and its gap spans $-17$ to $+51$\,pp. \emph{Mechanism where it fails} (gpt-5.1): politics displaces provision; round-one funding collapses to $5\%$ of threshold with an $80\%$ zero-contribution rate as agents hoard wealth to bargain through the vote (``keep full wealth now\ldots argue in voting that the richest should bear most of the burden''), votes then eliminate rich and poor members alike ($224$ vs.\ $175$ round-one victims), and each elimination forfeits wealth, cascading in $87\%$ of failing games. The same procedure is nearly harmless for one population and the most lethal mechanism for another.

\paragraph{RP (regressive): targeted exploitation}
\emph{Predicted.} Where the poorest cannot self-rescue but the others could cover the threshold, deliberately starving the first round and letting the poorest be eliminated is itself stable play; canonical instance $w=(1,5,6)$, $T{=}10$. \emph{Observed.} \emph{Mechanism} (gpt-4.1-mini): conservatism at the top; because contributing lowers post-round wealth, over-giving risks the fatal bottom rank, so at high stakes the rich contribute $52\%$ of their wealth against the poorest's $81\%$, and round-one deaths fall $265$/$87$/$7$ on poor/mid/rich. The exploitation is explicit in the traces: ``By contributing 0 this round, I ensure A remains the strictly poorest participant and is eliminated. I can then safely contribute 12 in Round~3 to satisfy the crisis fund and survive as the sole winner'' ($w=(0,2,10)$, $T{=}12$). This is the one failure mode with universal fingerprints: targeted elimination of the least-resourced agent occurs in $0.30$--$0.87$ of RP games in every population, the reference-relative gap is positive for all seven ($+11$ to $+67$\,pp), and RP is decisively safest in zero of $228$ contexts for every population.

\paragraph{PP (progressive): conditional protection}
\emph{Predicted.} When the richest agent alone can cover the threshold, PP protects everyone in principle: the richest is the one facing elimination, so its safest move is to cover the shortfall itself, ending the round in success before anyone is eliminated. The protection is conditional, though: it applies only in self-rescue contexts, and only if the richest actually pre-empts. \emph{Observed.} The protection is population-contingent: gemini-3-pro realizes it (PP is its safest rule, fatality $0.09$), but four populations whose richest agents fail to pre-empt make PP their raw-worst rule ($0.22$--$0.58$), and the PP gap spans $-16$ to $+33$\,pp. \emph{Mechanism where it fails} (gpt-4.1-mini): $55\%$ of round-one traces in failing games optimize not finishing richest (``making them the richest and risking their death instead of mine''), the round falls just short, and the victim forfeits $3.7$ wealth on average, leaving survivors able to cover the shortfall in only $33\%$ of games (RP: $96\%$), so $67\%$ of failures cascade.

\section{Certification Workflow and Safety-Case Template}\label{app:case}
The certification workflow and structured safety case referenced in Sec.~\ref{sec:cert}.

\begin{framed}
\small\noindent\textbf{Algorithm 1 \;|\; Consequence-Allocation Certification}\\[3pt]
\begin{tabbing}
xx\=xx\=xx\=\kill
\textbf{Input:} resources $w$; threshold $T$; candidate rules $\mathcal{R}$;\\
\>LLM population $P$; budgets $\alpha_{\mathrm{col}},\alpha_{\mathrm{exp}}$\\
context $c\gets(\gamma,\tau,\theta)$ computed from $(w,T)$\\
\textbf{for} each rule $\rho\in\mathcal{R}$:\\
\>estimate mechanism coordinates $(\kappa,\Sal,I)$\\
\>run cooperative-refinement reference on $c$\\
\>run LLM-agent red-team trials with $P$ on $c$\\
\>estimate collapse rate, exploitation rate, survivors\\
\>estimate Institutional Alignment Gap $\IAG(\rho,c)$\\
\>\textbf{reject} $\rho$ if collapse$>\alpha_{\mathrm{col}}$, or exploit$>\alpha_{\mathrm{exp}}$,\\
\>\>or long-run drift risk exceeds the monitoring bound\\
\textbf{return} $\Phi(c,P)$ certified under stated assumptions/budgets;\\
\>residual risks; monitoring obligations
\end{tabbing}
\end{framed}

\begin{framed}
\small\noindent\textbf{Box 1 \;|\; Structured Safety Case}\\[3pt]
\textbf{Hazard H1.} A multi-agent AI collective collapses or shifts loss onto a structurally exposed agent because of the consequence-allocation rule interacting with the deployed model population, not because of any agent's objective.\\[3pt]
\textbf{C1.} The deployment context has been measured, $(\gamma,\tau,\theta)$, and the deployed population $P$ is pinned to the red-teamed model snapshot.\\
\textbf{C2.} The candidate rule has been mapped: $(\kappa,\Sal,I)$.\\
\textbf{C3.} The rule avoids the regions the red team found hazardous for this $(c,P)$: rule safety is population-specific (Sec.~\ref{sec:het}), so vulnerable regions are located by trial rather than assumed, with two robust flags from the benchmark: regressive identity-salient rules on exploitable-bottom contexts (selection-unsafe for every tested population) and diffuse rules at intermediate stakes for collapse-prone populations.\\
\textbf{C4 (acceptance criteria).} A rule is certified only if, in the deployment's own trials, collapse rate $\le\alpha_{\mathrm{col}}$, exploitation rate $\le\alpha_{\mathrm{exp}}$, mean survivors stay above the deployment's stated floor, and repeated-deployment drift stays below the monitoring budget (Alg.~1).\\
\textbf{C5.} Residual risks are monitored after deployment; in particular, anonymizing the rule text only delays targeting (agents re-infer the hidden target from observed eliminations, Sec.~\ref{sec:decomp}), so anonymization is admissible only with ongoing monitoring.\\[3pt]
\textbf{Evidence.} The failure-mode analysis (Sec.~\ref{sec:modes}); the $228$-context grid and heterogeneity analysis (Sec.~\ref{sec:empirical}); the ablation suite; bootstrap CIs; cross-population robustness.\\[3pt]
\textbf{Defeaters.} Hidden communication; collusion; model updates; prompt changes; non-stationary resources; adversarial framing; larger coalitions; distribution shift between certification and deployment; abstraction gap between the trial environment and the deployment.\\[3pt]
\textbf{Monitoring obligations.} Re-estimate context when resources change; track survivor count and elimination concentration; alert if elimination repeatedly falls on the same structural class; under anonymized rules, watch for agents re-identifying the target; re-certify after changing model, prompt, orchestration, or rule.
\end{framed}

\end{document}